# Comparison of Stereo Matching Algorithms for the Development of Disparity Map

Hamid Fsian[a*], Vahid Mohammadi[a], Pierre Gouton[a], Saeid Minaei[b]

[a] ImViA, UFR Sciences et Techniques, Université de Bourgogne, Franche-Comté, Dijon, France
[b] Biosystems Engineering Department, Faculty of Agriculture, Tarbiat Modares University, Tehran, Iran

* Corresponding author. Tel: +33-751986711

Email address: Abdelhamid-Nour-Eddine.Fsian@u-bourgogne.fr, abdelhamidfsian@gmail.com

Postal address : 87 Route de Beaune, 21000, Dijon, France

## Abstract

Stereo Matching is one of the classical problems in computer vision for the extraction of 3D information but still controversial for accuracy and processing costs. The use of matching techniques and cost functions is crucial in the development of the disparity map. This paper presents a comparative study of six different stereo matching algorithms including Block Matching (BM), Block Matching with Dynamic Programming (BMDP), Belief Propagation (BP), Gradient Feature Matching (GF), Histogram of Oriented Gradient (HOG), and the proposed method. Also three cost functions namely Mean Squared Error (MSE), Sum of Absolute Differences (SAD), Normalized Cross-Correlation (NCC) were used and compared. The stereo images used in this study were from the Middlebury Stereo Datasets provided with perfect and imperfect calibrations. Results show that the selection of matching function is quite important and also depends on the images properties. Results showed that the BP algorithm in most cases provided better results getting accuracies over 95%.

# 1 Introduction

Stereo matching is a fundamental topic in computer vision systems in which two cameras from different viewpoints are utilized to extract 3D information by evaluating the relative positions of objects in two perspectives of the scene. Stereo matching is the process of determining relative displacements between image pairs captured by stereo cameras (Egnal, 2000; Scharstein et al., 2014; Kim, 2003). Despite the smaller search space, discovering stereo correspondences in real-world images remains complicated to occlusions, reflective surfaces, repeated patterns, textureless or low-detail regions that can alter the similarity measure and hinder the search. The similarity of image locations is measured in all stereo correspondence techniques. For all disparities under consideration, a matching cost is typically estimated at each pixel. On gray and color images, common pixel-based matching costs include absolute differences, squared differences, sampling-insensitive absolute differences, or reduced versions (Dalal and Triggs, 2005). The Sum of Absolute or Squared Differences (SAD / SSD), Normalized Cross-Correlation (NCC), and rank and census transforms are all common window-based matching costs (Mikolajczyk and Schmid, 2005). Filters can be used to efficiently implement some window-based expenses. For example, a rank filter followed by absolute differences of the filter results can be used to compute the rank transform. Other filters, such as LoG and mean filters, attempt to reduce bias or gain changes in the same way. Mutual information (Scharstein et al., 2014) and approximatively segment-wise mutual information, as utilized in stereo method (Hirschmuller and Scharstein, 2007), are two more sophisticated similarity measures.

On the other hand, other algorithms are commonly used like the Belief propagation. Despite these significant advancements, belief propagation systems still take a long while even on strong fastest desktop computers to process for solving stereo problems. When it comes to optical flow and picture restoration concerns, they are too slow to be useful. As a result, one must choose between these methods, which yield good results but are time-consuming, and local methods, which yield significantly lower results but are quick. However, these algorithms have been improved (Felzenszwalb and Huttenlocher, 2006), to give great results with less computing time.

This study also takes a look at a feature-based stereo correspondence technique, focusing on the Histogram of Oriented Gradients (HOG), Gradient Features (GF), and Block Matching (BM) approach and BM with Dynamic Programming for determining depth in 2-D images. HOG is a feature-descriptor used in computer vision to detect objects by counting the number of occurrences of gradient orientation in a specific region of an image. This approach is similar to other approaches such as edge oriented histogram, speeded up robust features (SURF), and scale-invariant feature (SIFT) descriptor as they are all feature descriptors. However, there have been criticisms of the feature-based methods method, as it is highly dependent on the window size used for correspondence, which is fixed throughout the entire algorithm, and there are some merits, such as the fact that features extraction is mostly invulnerable to shadowing and illumination (Aboali et al., 2017). There has been performed much research on the use of different matching algorithms for stereo image processing (Chai and Cao, 2018; Mozerov and Weijer, 2015; Heise et al., 2013; Pham and Jeon, 2012; Çiğla and Alatan, 2011; Geiger et al., 2010). Gong et al. (2007) studied the careful selection and aggregation of matching costs of neighboring pixels. Six cost aggregation approaches were examined and reported that properly designed cost aggregation approaches remarkably increased the quality of the disparity maps. Zhu and Yan (2017) adopt a modified Census transform with a local texture metric and then aggregated the costs with guided image filter. It was observed that their proposed method achieved an average error rate of 5.22 % on the Middlebury dataset.

This paper helps choosing the most suitable algorithm to estimate disparity knowing the characteristics of the image pairs. Different matching algorithms and cost functions were applied on three categories of images based on their details and texture features.

## 2 Materials and methods

### 2.1 Datasets of stereo images

In this study, two datasets from the Middlebury Stereo Vision database were used which consist of high-resolution image pairs and their ground truth. In addition, for each scene images with perfect and imperfect calibration have been provided. Table 1 presents the details of the datasets used in this study. Middlebury Stereo Vision dataset consists of a structured lighting technique for producing high-resolution stereo datasets of static interior scenes with extremely realistic ground-truth discrepancies. Figure 1 represents the left images of all image pairs considered for the development of disparity map.

**Table 1.** The databases of stereo and ground truth images used in this study.

| Source No. | Source | Link to the database | Number of image pairs | Reference |
|---|---|---|---|---|

| 1 | Middlebury | *https://vision.middlebury.edu/stereo/data/scenes2014/* | 33 | *Scharstein* |
| 2 | Stereo vision | *https://vision.middlebury.edu/stereo/data/scenes2021/* | 24 | *et al. 2014* |

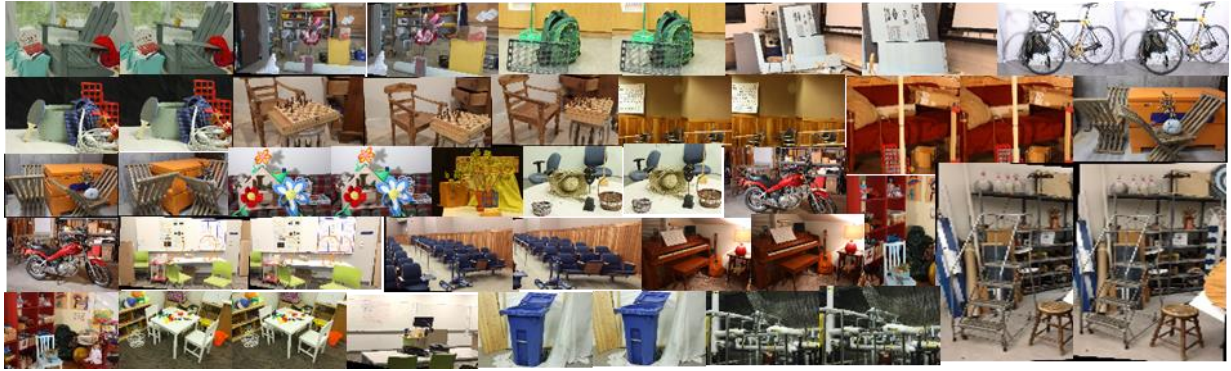

**Fig. 1.** Total of left images used in this study provided by the Middlebury Stereo Dataset.

**2.2 Stereo Algorithms**

The algorithm that employs the cost can affect the performance of a matching cost. Thus six different algorithms have been considered which are including Block Matching (BM), Block Matching with Dynamic Programming (BMDP), Belief Propagation (BP), Gradient Feature Matching (GF), Histogram of Oriented Gradient (HOG), and the proposed method where the cost function is aggregated with fixed window sizes. The SAD cost function was utilized for the BM, BMDP and the aggregated method. The MSE was implemented on the BM and BMDP and the NCC cost function in the aggregated method.

*2.2.1 Block Matching Algorithm*

BM consists of Sliding a window along the epipolar line and compares contents of that window with the reference window in the left image (Aboali et al., 2017). Then a cost function will compute the difference between those two blocks, and finally the block with most similarities with the reference block will be matched. Following this idea, the most crucial point in this algorithm is the choice of the block size, regardless to the cost function. In fact, if a small window size is chosen, there will be more details but also more of noise. On the other hand, a larger window size implies a smoother disparity map with less details and fails near boundaries.

*2.2.2 Block Matching with Dynamic Programming*

This algorithm is used in order to avoid situation where BM fails to give a good result, especially when image pairs are textureless regions, repeated patterns and specularities. In this paper, the dynamic programming method was chosen as the global optimization technique for the disparity optimization phase since this algorithm optimizes the energy function to be NP-hard(non-deterministic polynomial-time hardness) for the purpose of smoothness and enhancement.

*2.2.3 Belief Propagation*

BP method is set to perform inference on Markov random fields (Felzenszwalb and Huttenlocher, 2006). The max-product approach, in particular, can be used to approximate minimum cost labeling of energy functions. This algorithm is normally specified in terms of probability distributions, but it can also be computed with negative log probabilities, in which case the max-product becomes a min-sum. The choice of this formulation is due the low numerical errors and it directly employs the energy function concept. The max-product BP algorithm operates by sending messages around the four-connected picture grid's graph. Each message is a vector with a dimension equal to the number of labels that can be used.

*2.2.4 Histogram of Oriented Gradient*

The idea behind HOG Descriptor (Image descriptors are descriptions of the visual features of the contents in images, videos, or algorithms or applications that produce such descriptions. They describe elementary characteristics such as the shape, the color, the texture or the motion, among others.) is that edge direction and intensity gradients can be used to characterize an object's form and appearance in an image (Dalal and Triggs, 2005). However, before we can calculate the histogram, we must first partition the image into smaller connected sections of a defined size. Only then a histogram of gradient orientations can be computed, but not for the full image, rather for each individual cell, resulting in several histograms equal to the number of accessible cells, which add up to the required descriptor. It's also worth noting that orientation can be represented as angles between [0, 180] unsigned and [0, 360], with the choice of (un)signed depending on the project and the required degree of gradient accuracy.

Normalizing-by-contrast is a common optimization for intensity-based descriptors in which we compute the intensity over a bigger region termed block and then utilize this newly-found intensity information to normalize all the cells within the block.

**2.3 Matching Cost**

Our initial cost function is the widely used absolute difference, which assumes brightness consistency for corresponding pixels and acts as our evaluation's baseline performance metric. Local stereo methods commonly SAD behave over a window, whereas global stereo methods use pixel-by-pixel differences (Equation 1).

$$SAD = \sum_{(i,j) \in W} |I_1(i,j) - I_2(x+i, y+j)| \tag{1}$$

where $I_1$ refers to the reference image, while $I_2$ indicates the target image, and $W$ indicates the square window for aggregation.

Another cost function that was used is the Mean Squared Error (MSE) (Equation 2). For images, the 'error' in MSE is a synonym of 'difference'. Then a difference between two image pairs is required to be obtained, but also the 'mean' is a synonym of 'average'. The sum of differences is needed to be divided by the number of the pixel. So, it can be called Average Squared Difference.

$$MSE = \frac{1}{MN} \sum_{n=1}^{M} \sum_{m=1}^{N} [\hat{g}(n,m) - g(n,m)]^2 \tag{2}$$

Where $g$ refers to the reference image, while $\hat{g}$ indicates to the target image.

Next, Normalized Cross-Correlation (NCC) (Equation 3) was the other cost function used for error measurement. NCC is a method for matching two windows around a pixel of interest that is widely used. The window's normalization accounts for variances in gain and bias. The statistically best method for correcting Gaussian noise is NCC. However, because outliers create substantial errors in the NCC computation, NCC tends to obscure depth discontinuities more than many other matching costs. Moravec first proposed MNCC as a common variation (Moravec, 1977). The standard NCC was chosen because MNCC produced inferior outcomes. Where $I1$ refers to the reference image, while $I2$ indicates to the target image, W indicates the square window for aggregation.

$$NCC = \frac{\Sigma_{(i,j)} I_1(i,j) \cdot I_2(x+i, y+j)}{\sqrt[2]{\Sigma_{(i,j)} I_1^2(i,j) \cdot \Sigma_{(i,j)} I_2^2(x+i, y+j)}} \tag{3}$$

## 2.4 Programming

The flow chart below shows how the complete algorithm works. The left and right images and the ground truth are read and sent to preprocessing. During this step, the RGB images were transformed into gray images and then resized. The resizing was done to reduce the processing time. The format of the images from integer-eight was changed to double. Next, the images were presented to the disparity matching algorithm. All the programs were developed using MATLAB software (MathWorks, 2019b, the USA).

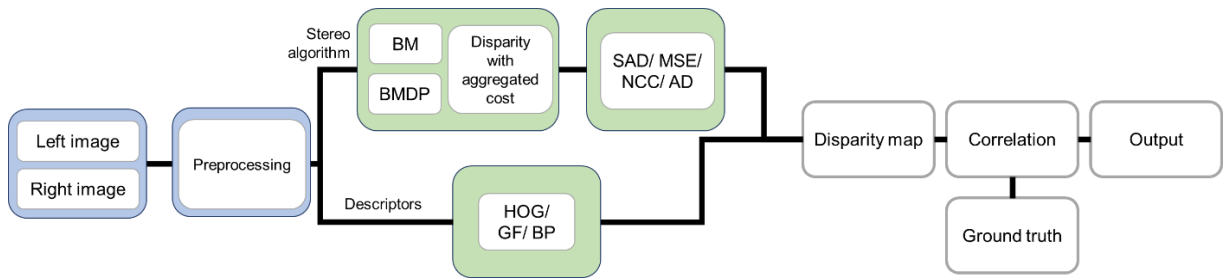

**Fig. 2.** Diagram showing the course of the whole algorithms

## 4 Results and Discussion

Stereo matching has been a challenging task and many algorithms have been developed to optimize it. However, the choice of matching techniques and functions play an important role for having promising results. This work compares several cost functions that are used for stereo image matching. After computing the disparity map for all transformations and all combinations of matching costs and stereo techniques, the analysis of the result was made by computing correlation between the disparity map and the ground truth provided by the Middlebury stereo datasets.

### 4.1 SAD

The results of matching techniques with the SAD cost function were pretty different. As Figure 3 shows, the BP provides the best result for the matching. However, it is observed that still the result depends on the image and the details inside. For example in the image of the motorcycle (Fig. 4), the DWAC had the best matching. Also, it is seen that the HOG technique is so sensitive to the details and in Fig. 4 the details have been detected much more neatly.

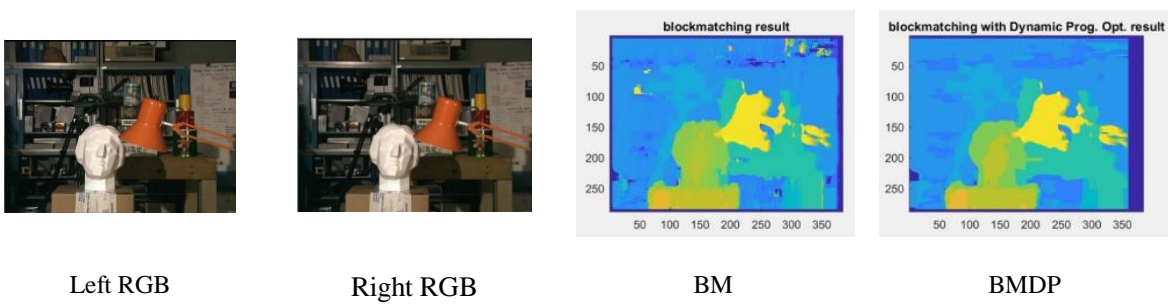

Left RGB      Right RGB      BM      BMDP

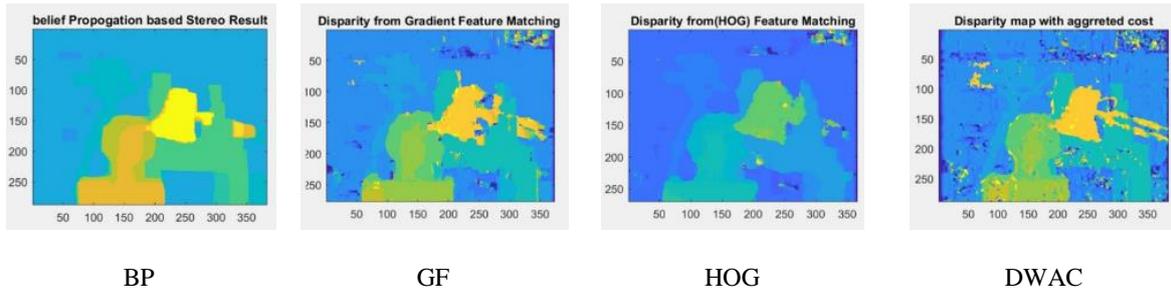

| BP | GF | HOG | DWAC |

**Fig. 3.** Left and right images and the disparity maps achieved for different matching algorithms based on SAD cost function.

## 4.2 MSE

Figure 5 provides the disparity maps that constructed using MSE cost function for different matching algorithms. The results are quite similar to the SAD algorithm. However, the disparity maps with MSE also are proper for BP, GF and HOG and for the rest the results are not very acceptable. Figure 6 provides an example of a difficult task of matching as the images contain less details and textures.

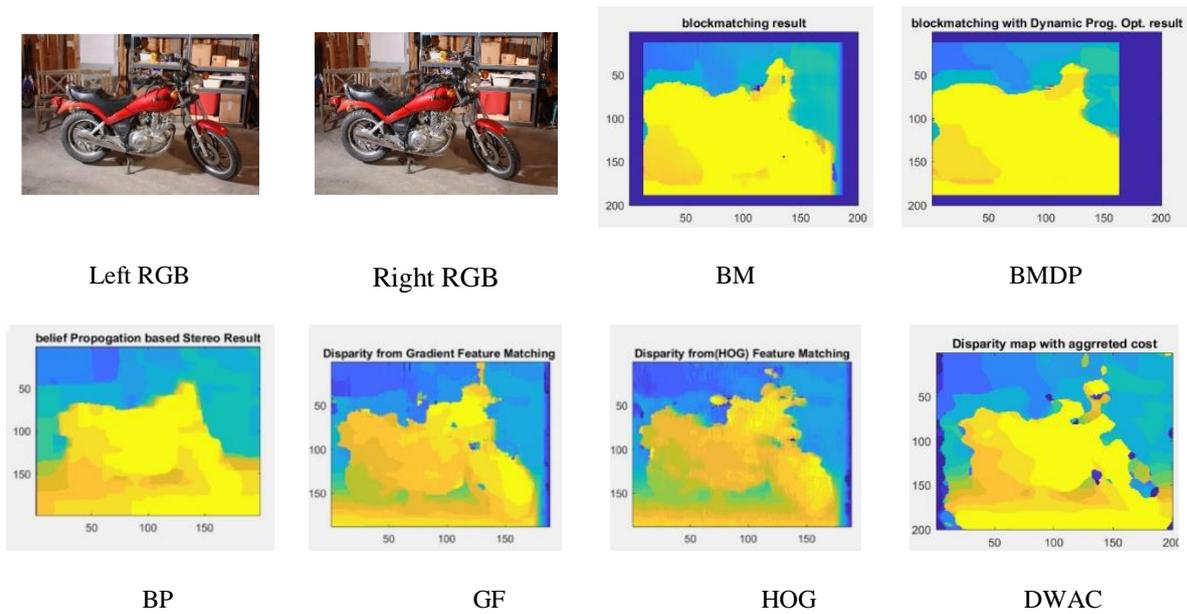

| Left RGB | Right RGB | BM | BMDP |

| BP | GF | HOG | DWAC |

## 4.3 NCC

Hereafter, the results obtained by NCC are presented. As figures 6.44-46 show, the performance of NCC was not as good as MSE and SAD for different matching algorithms. Here also the performance of HOG has been more precise and clear. Hirschmuller and Scharstein (2007) reported the same results. They found NCC the least interesting results while other techniques were much better.

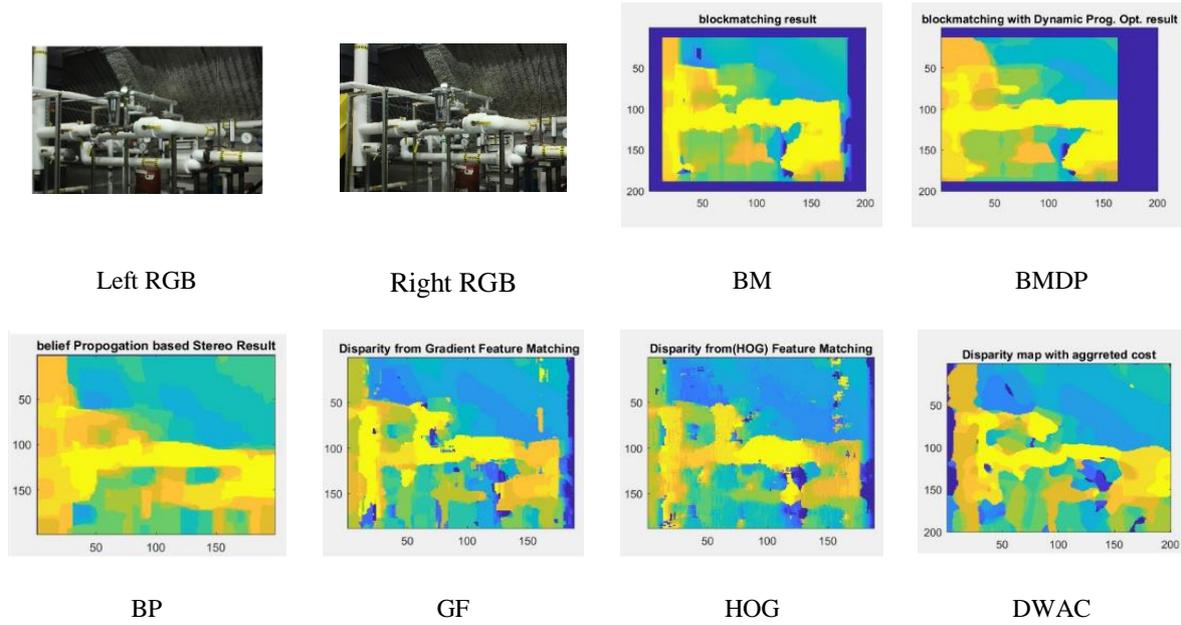

| | | | | | | | Left RGB | | | Right RGB | | BM | | BMDP |
|---|---|---|---|---|---|---|---|---|---|---|---|---|---|---|

| BP | | GF | | HOG | | DWAC |

**Fig. 5.** NCC for motorcycle, curule1 and Pipes

Table 3 summarizes the minimum and maximum of average errors for all images used for disparity map development using different matching and cost algorithms. As the table indicates, the error strongly depends on the type of application, details, and textures inside the image. The best result was obtained by the use of BP and SAD. Heo et al. (2008) proposed Adaptive Normalized Cross Correlation for stereo matching that performed well in correspondence. Among recent techniques is the use of deep learning for stereo matching. Huang et al. (2021) used PSMNet for stereo image matching for KITTI 2012 and KITTI 2015 datasets representing good results.

**Table 3.** Average results using correlation of all disparity maps with GT

| Method | BM | | BMDP | | BP | | GF | | HOG | | DWAC | |
|---|---|---|---|---|---|---|---|---|---|---|---|---|
| | Min | Max | Min | Max | Min | Max | Min | Max | Min | Max | Min | Max |
| MSE | 0.113 | 0.853 | 0.106 | 0.903 | 0.272 | 0.933 | 0.099 | 0.854 | 0.128 | 0.905 | 0.043 | 0.841 |
| NCC | 0.236 | 0.881 | 0.156 | 0.925 | 0.315 | 0.941 | 0.153 | 0.904 | 0.246 | 0.91 | 0.121 | 0.892 |
| SAD | 0.291 | 0.919 | 0.253 | 0.94 | 0.347 | 0.964 | 0.21 | 0.919 | 0.295 | 0.94 | 0.102 | 0.904 |

Figure 7 shows the stereo matching accuracy for all image pairs used in this study. It is observed that BP has had mostly the best accuracies, however for all algorithms the accuracy of matching has not been fixed and has experienced fluctuations for different types of images.
4

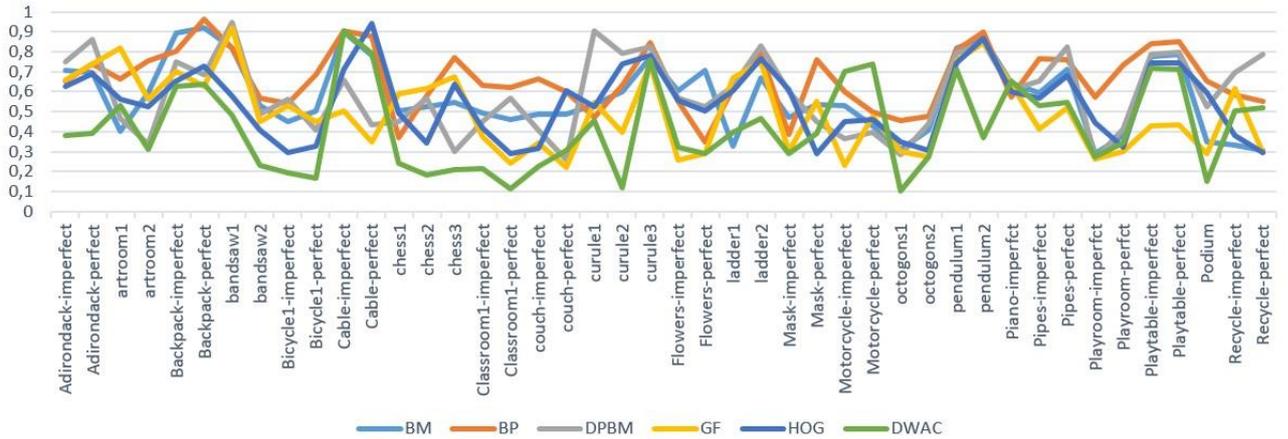

**Fig. 7.** The whole comparison of all algorithms for all the images.

## 5 Conclusion

The analyses reveal that the global optimization strategy (BP), which considers neighborhood disparity on both the horizontal and vertical axes, outperforms the others by a significant margin. Because the simple block matching approach just tries to identify the lowest SAD, MSE and NCC value regardless of pixel neighborhood, it fails in a large number of pixels, especially when noise is included. However, while dynamic programming can manage some errors, it can't handle all of them. It produces superior horizontal results, as you can see. However, because it ignores the vertical neighbors, a large number of pixels are lost. In the other hand, Descriptor methods like HOG and GF done well in comparison of the BP algorithm, but the computation time where huge. Finally, the belief propagation strategy provides us with superior outcomes in both time and performances.

**Figures captions**

**Fig. 1.** Flow chart showing the course of the whole algorithms

**Fig. 2.** Flow chart of Preprocessing steps

**Fig. 3.** SAD for tsubaka, motorcycle and chess3

**Fig. 4.** MSE for motorcycle, Adirondack and artroom1

**Fig. 5.** NCC for motorcycle, curule1 and Pipes

**Fig. 6.** Comparison of all Algorithms with correlation

**Tables captions**

**Table 1.** The databases of stereo and ground truth images used in this study.

**Table 2.** Comparison between cost functions.